\documentclass[10pt,twocolumn,letterpaper]{article}

\usepackage{tabularx}

\usepackage{cvpr}              

\definecolor{cvprblue}{rgb}{0.21,0.49,0.74}
\usepackage{subcaption} 
\usepackage[pagebackref,breaklinks,colorlinks,allcolors=cvprblue]{hyperref}
\usepackage[T1]{fontenc}

\def\paperID{6824} 
\def\confName{CVPR}
\def\confYear{2025}

\title{4Real-Video: Learning Generalizable Photo-Realistic 4D Video Diffusion}

\author{Chaoyang Wang\textsuperscript{1,}\footnotemark[1] \footnotemark[2]
\hspace*{1em} 
Peiye Zhuang\textsuperscript{1,}\footnotemark[1]
\hspace*{1em} 
Tuan Duc Ngo\textsuperscript{1,2}
\hspace*{1em} 
Willi Menapace\textsuperscript{1}
\hspace*{1em} 
Aliaksandr Siarohin\textsuperscript{1}
\\ 
Michael Vasilkovsky\textsuperscript{1}
\hspace*{1em} 
Ivan Skorokhodov\textsuperscript{1}
\hspace*{1em} 
Sergey Tulyakov\textsuperscript{1}
\hspace*{1em} 
Peter Wonka\textsuperscript{1,3}
\hspace*{1em} 
Hsin-Ying Lee\textsuperscript{1}
\\
\textsuperscript{1}Snap Inc \, \textsuperscript{2}Umass Amherst  \,
\textsuperscript{3}KAUST\\
\url{https://snap-research.github.io/4Real-Video/}
}

\def\Approach{4Real-Video\xspace}

\begin{document}

\twocolumn[{%
	\renewcommand\twocolumn[1][]{#1}%
	\maketitle
        \vspace{-10mm}
	\begin{center}
		\includegraphics[width=\linewidth]{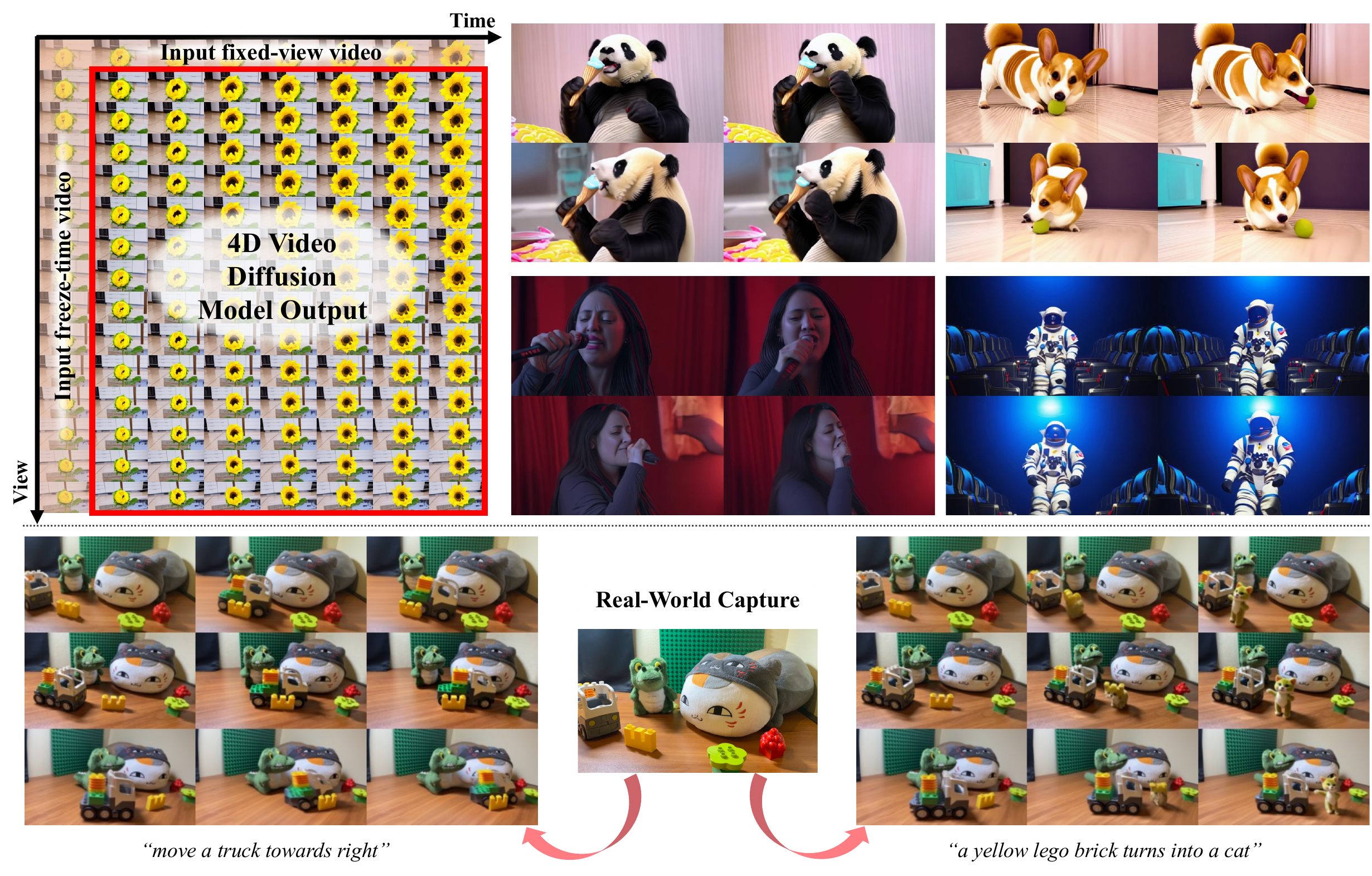}
            \vspace{-8mm}
		\captionof{figure}{
        \textbf{\Approach} is a 4D generation framework that (\textit{top-left}) takes a fixed-view video and a freeze-time video as input and generates a grid of consistent video frames. One axis of the grid varies in time, and the other axis varies the viewpoint. The input videos can be real videos or videos generated by a video model. Note that our method can generate grids larger than $8 \times 8$ videos. Here, we present subsets of frames as an example.
        (\textit{top-right}) 4D videos generated from generated videos.
        (\textit{bottom}) We can also capture a real-world scene, and generate a 4D video given different prompts.}
        \vspace{-1mm}
		\label{fig:teaser}
	\end{center}    
}]
\renewcommand{\thefootnote}{} 
\footnotetext{* main contributor, $\dagger$ project lead.}
\renewcommand{\thefootnote}{\arabic{footnote}} 

\maketitle

\begin{abstract}
We propose \Approach, a novel framework for generating 4D videos, organized as a grid of video frames with both time and viewpoint axes.  In this grid, each row contains frames sharing the same timestep, while each column contains frames from the same viewpoint.
We propose a novel two-stream architecture.
One stream performs viewpoint updates on columns, and the other stream performs temporal updates on rows. After each diffusion transformer layer, a synchronization layer exchanges information between the two token streams. We propose two implementations of the synchronization layer, using either hard or soft synchronization.
This feedforward architecture improves upon previous work in three ways:
higher inference speed, enhanced visual quality (measured by FVD, CLIP, and VideoScore), and improved temporal and viewpoint consistency (measured by VideoScore and Dust3R-Confidence). 
\end{abstract}    
\section{Introduction}
\label{sec:intro}
\vspace{-1mm}
With the recent rise of video diffusion models~\cite{sora,moviegen,cogvideox}, \emph{4D video} generation has emerged as an important extension. \emph{4D video} generation has numerous potential applications, including creating dynamic scenes and objects through post-processing and enabling immersive experiences via image-based rendering techniques. 
To position our work, we define \emph{4D video} as follows: 4D video is a grid of video frames with a time and a view-point axis. 
In our arrangement, all frames in a row share a timestamp, and all in a column share a viewpoint (see Fig.~\ref{fig:teaser}, left for an example).
Our definition contrasts with recent work that also uses the term ``4D video", to describe video generation with camera and motion control~\cite{genxd}. To clarify this distinction, we will refer to such approaches as \emph{camera-aware}.
While both paradigms share similar applications, we believe that \emph{4D video} has two important advantages compared to \emph{camera-aware video}: (a) a complete space-time grid can provide full 4D experiences and enable easier dynamic reconstruction, yet it is non-trivial for camera-aware methods to complete such a grid, and (b) videos generated by camera-aware methods tend to have inferior multi-view consistency~\cite{cvd}.

As 4D video generation is a very recent topic, there are only a few competing approaches. 
Some works~\cite{sv4d,vividzoo,4diffusion} propose training the 4D models directly using the limited available 4D data, such as synthetic animated 3D assets from Objaverse~\cite{objaverse} or a human-specific dataset~\cite{human4dit}.
These models can generate a space-time grid, yet they cannot generalize beyond the limited training data distribution.
Furthermore, the architecture designs, which sequentially interleave temporal and view attention, often fail to account for their interdependence, leading to artifacts or reduced generalization.

To address the challenges of generating 4D videos, we introduce a novel multi-view video generation model leveraging a two-stream architecture to enhance multi-view and temporal consistency. 
Our approach extends existing transformer-based video diffusion models by splitting video tokens into two streams: one dedicated to capturing temporal updates across fixed viewpoints and the other focused on view updates across freeze-time frames.
These streams are processed independently using pre-trained transformer layers to reuse existing models efficiently. 
To ensure coherence between the streams, we introduce a synchronization layer that dynamically exchanges information between the temporal and view tokens. 
Inspired by the optimization literature, we propose two types of synchronization layers that perform either hard or soft synchronization updates, with the latter providing greater flexibility by learning adaptive weights to modulate token interactions across layers. This design avoids the distributional shifts observed in sequential model designs, preserves the consistency of the original video model, and enables high-quality 4D generation.

The proposed architecture design can generate diverse, dynamic multi-view videos in approximately 1 minute ($8\times 8$ frames at a resolution of $288 \times 512$), as opposed to hours required by previous SDS-based approaches~\cite{4real,4dfy,dreamgaussian4d,ayg}. 
Beyond its speed advantage, the model generalizes well with limited 4D training data. This is achieved by initially training on 2D transformed videos to simulate synchronized camera motion, followed by fine-tuning on a small amount of animated Objaverse data~\cite{objaverse}.
Moreover, our model does not rely on explicit camera conditioning modules. Instead, it takes a real or generated freeze-time video and a fixed-view video as conditional inputs, automatically inferring the viewpoints and motion to be generated. This effectively decomposes the problem, allowing us to leverage recent advancements in camera-controlled video generation as conditional input. It also simplifies the process of animating existing freeze-time videos by removing the requirement for users to provide camera poses explicitly.

In summary, we make the following contributions:
\begin{itemize}
\item We propose a two-stream architecture for 4D video generation that independently handles temporal and view updates, synchronizing streams to ensure consistency.
\item We propose flexible synchronization mechanisms that enable efficient and adaptive token interactions, preserving the generation quality of pre-trained video layers.
\item Our model is data-efficient and can produce high-resolution 4D videos in a fraction of the time required by prior methods. We obtain state-of-the-art results in terms of video quality and multi-view consistency.
\end{itemize}

\vspace{-1.5mm}
\section{Related Work}
\label{sec:related_work}
\vspace{-1.5mm}

\paragraph{Optimization-Based 4D Generation.}
Score Distillation Sampling (SDS)~\cite{dreamfusion,prolificdreamer,sjc,fantasia3d,magic3d,hifa} has been used to generate 3D content by obtaining gradients from pre-trained models like text-to-image~\cite{ldm,imagen} and text-to-multiview models~\cite{zero123,mvdream}.
Extending this approach, a branch of 4D generation methods~\cite{4dfy,ayg,consistent4d,dreamgaussian4d,4dgen,animate124,mav3d} leverages additional text-to-video supervision~\cite{animatediff,imagenvideo,videocrafter} to generate dynamic content.
However, these methods require time-consuming optimization processes, often requiring hours to produce a 4D output.
Furthermore, most methods derive 3D priors from multi-view diffusion models~\cite{zero123,mvdream} trained on an object-centric and synthetic dataset~\cite{objaverse}, resulting in a bias toward object-centric, non-photorealistic outputs.

\begin{figure*}[t]
\begin{center}
\includegraphics[width=\linewidth]{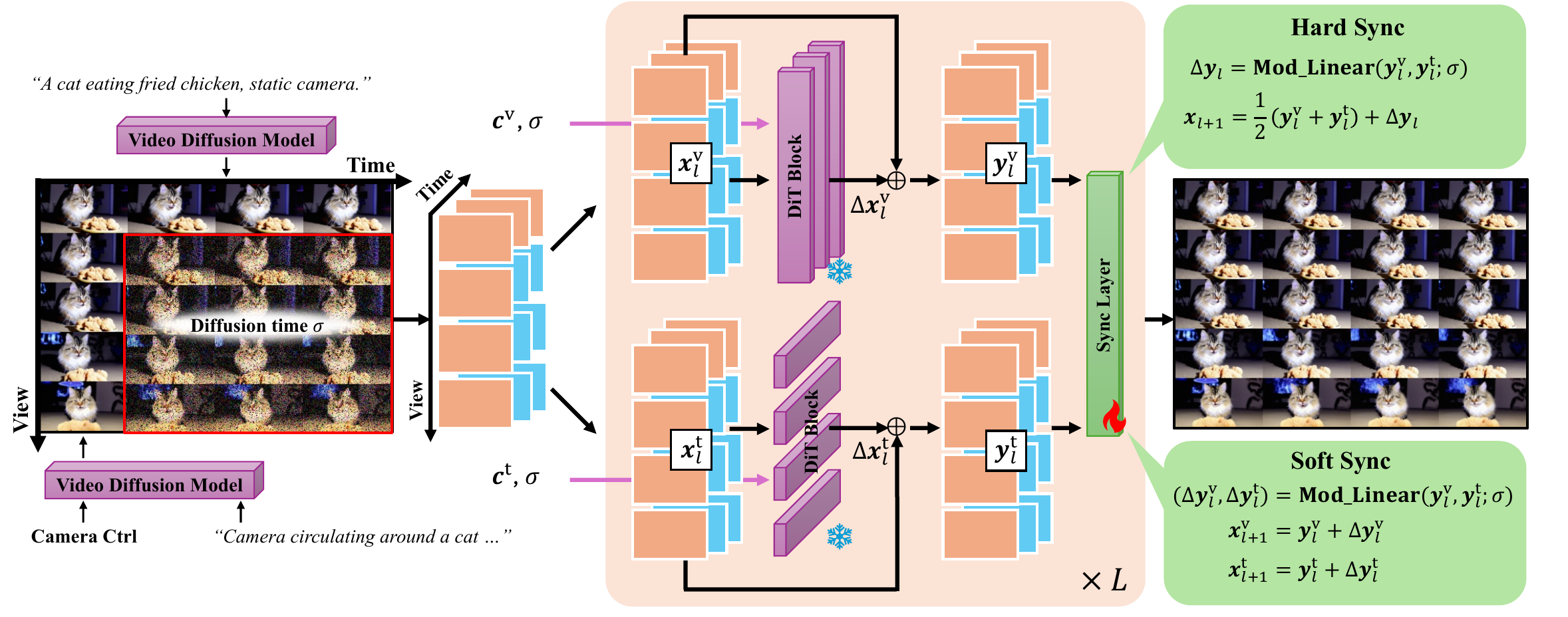}
\vspace{-5mm}
\caption{\textbf{Overview of \Approach.} Left: we initialize the grid of frames with a (generated or real) fixed-viewpoint video in the first row and a freeze-time video in the first column. Middle: our architecture consists of two parallel token streams. The top part processes $\mathbf{x}_l^\text{v}$ with viewpoint updates and the bottom part processes $\mathbf{x}_l^\text{t}$ with temporal updates. Subsequently, a synchronization layer computes the new tokens $\mathbf{x}_{l+1}^\text{v}$ $\mathbf{x}_{l+1}^\text{t}$ for the next layer in the diffusion transformer architecture. Right: we propose two implementations of the synchronization layer: hard and soft synchronization.
}
\label{fig:overview}
\end{center}
\vspace{-6mm}
\end{figure*}

\noindent\textbf{Camera-aware video generation}
Text-to-video models~\cite{sora,cogvideox,moviegen,snapvideo} have shown promising results in generating coherent and photorealistic video content. 
To enable a more controllable and interactive content creation process, camera control in video generation has gained attention.
These approaches~\cite{vd3d,motionctrl,cameractrl,directavideo,4DiM} propose adding camera control by injecting camera pose information into the temporal layers. 
Additionally, some approaches~\cite{genxd,camco} collect and annotate videos with camera poses to fine-tune video models.
However, despite being visually consistent, the content generated by camera-aware methods tends to have multi-view inconsistencies. 
Furthermore, it is not trivial for camera-aware video models to generate a complete space-time grid of 4D videos, limiting their applicability in fully 4D scenarios.

\noindent\textbf{4D video generation}
In this work, we define \textit{4D video} as a video grid organized along both temporal and viewpoint axes.
Some work~\cite{sv4d,vividzoo,4diffusion} trains 4D models using existing 4D data, typically subsets of Objaverse~\cite{objaverse}.
Although these models are conceptually capable of moving beyond object-centric content, their outputs remain constrained by the limited diversity of available 4D datasets in practice.
To reduce reliance on synthetic data, alternatives like 
4Real~\cite{4real} relies solely on a video model to generate consistent dynamic and freeze-time videos, followed by an optimization-based 4D reconstruction to obtain underlying 4D contents. 
However, the dynamic and freeze-time videos alone cannot guarantee consistency within the 4D grid, and the optimization is computationally expensive. 
CVD~\cite{cvd} tackles this limitation by fine-tuning video models to simultaneously generate structurally consistent video pairs, using pseudo-paired datasets curated from monocular video datasets~\cite{webvid10m,realestate10k}.
Although CVD proposes strategies to extend generation to multiple views, its consistency and efficiency remains suboptimal for multi-view 4D generation.

\section{Method}
\label{sec:method}

\noindent\textbf{Problem setup.}
We aim to generate a structured grid of video frames $\{I_{ij}\}$, where all frames in a row share a viewpoint, and all frames in a column share a timestep. In other words, each row is a fixed-view video, and each column is a freeze-time video.
The inputs to our method are the first row $I_{1*}$ and the first column $I_{*1}$ of the frames. These inputs are from either real-world videos or synthetic outputs from existing video generation models. The task is to synthesize the remaining frames while ensuring both temporal and multi-view consistency (see Fig.~\ref{fig:overview} left).

\vspace{-1mm}
\subsection{Base video model training}
\vspace{-1mm}
\noindent\textbf{Freeze-time and dynamic video generation.}
Inspired by 4Real~\cite{4real}, we train a base video model to support two distinct generation modes:
\emph{freeze-time video} that depicts static scenes with changes in viewpoint, and \emph{dynamic video} that captures object motion.
We group datasets into two categories: (1) videos with arbitrary camera and scene motions, and (2) videos of static scenes. Each group is associated with a unique context embedding that controls the generation process to align with the respective distributions. 

\noindent\textbf{Masked training.} 
To handle flexible input configurations, the model is trained using a random masking strategy. This enables the model to predict unseen frames based on any subset of input frames.
The design (1) allows for autoregressive generation of long videos by progressively synthesizing frames, and (2) provides essential flexibility for the 4D video model to condition on various input video frames.

\subsection{Multi-view video model}
\subsubsection{Two-stream architecture} 
Current state-of-the-art video diffusion models~\cite{sora,cogvideox,moviegen} mostly leverage a transformer-based architecture such as DiT~\cite{Peebles2022DiT}, which forwards video tokens through a series of spatial-temporal transformer blocks with skip connections. Specifically, each DiT transformer block $\varphi_l$ produces an update $\Delta\mathbf{x}_l$ to the current video tokens $\mathbf{x}_l$ at the $l$-th layer with condition $\textbf{c}$:
\vspace{-2mm}
\begin{equation}
        \Delta \mathbf{x}_l  = \varphi_l(\mathbf{x}_l;\mathbf{c}), \quad
        \mathbf{x}_{l+1}  = \mathbf{x}_{l} + \Delta \mathbf{x}_l.
\vspace{-2mm}
\end{equation}

In our setting, we need to extend to a set of tokens describing all frames in the grid. We use $\mathbf{x}_{l}$ to denote the set of all tokens at layer $l$ and $\mathbf{x}_{l,i,j}$ to describe the set of tokens for a frame at layer $l$ with time stamp $i$ and viewpoint $j$. As our goal is to reuse pre-trained high-quality video models as much as possible, we can utilize pre-trained DiT transformer layers to either update a row for view-point $i$ (Eq.~\ref{Eq:RowUpdate}) or a column for timestep $j$ (Eq.~\ref{Eq:ColumnUpdate}) of our frame grid,
\vspace{-1mm}
\begin{equation}    
\label{Eq:RowUpdate}
    \varphi_l^\text{v}(\{ \mathbf{x}_{l,i,1}, ..., \mathbf{x}_{l,i,T} \} ;\mathbf{c}) \quad \text{for } 1\leq i \leq V,
    \vspace{-1mm}
\end{equation}
\begin{equation}
\label{Eq:ColumnUpdate}
    \varphi_l^\text{t}(\{ \mathbf{x}_{l,1,j}, ..., \mathbf{x}_{l,V,j} \} ;\mathbf{c})  \quad \text{for } 1\leq j \leq Y.
    \vspace{-0.5mm}
\end{equation}
Since we have a total of $T$ timesteps and $V$ viewpoints, we can process the complete grid by either performing $V$ row updates or $T$ column updates in parallel when reusing existing DiT transformer blocks. To avoid overly complex notation, we write $\varphi_l^\text{v}(\mathbf{x}_{l},\mathbf{c})$ to denote the update of a single row or a parallel update of all $V$ rows jointly (and use  analogous 
 notation for column updates with $\varphi_l^\text{t}(\mathbf{x}_{l},\mathbf{c})$).

Our first important design idea is variable (token) splitting to create two separate sets of tokens to encode the complete frame grid, $\mathbf{x}_l^\text{t}$ for temporal and $\mathbf{x}_l^\text{v}$ for view updates. The set $\mathbf{x}_l^\text{v}$ will be processed using $T$ parallel row updates and the set $\mathbf{x}_l^\text{t}$ will be processed using $V$ parallel column updates. Updates are computed independently and in parallel:
\vspace{-0.5mm}
\begin{equation}
\mathbf{y}_l^\text{v} = \mathbf{x}_l^\text{v} + \varphi_l^\text{v}(\mathbf{x}_l^\text{v}; \mathbf{c}^\text{v}); \quad \mathbf{y}_l^\text{t} = \mathbf{x}_l^\text{t} + \varphi_l^\text{t}(\mathbf{x}_l^\text{t}; \mathbf{c}^\text{t}).
\vspace{-0.5mm}
\label{eq:parallel}
\end{equation}

We propose a synchronization layer after each DiT transformer layer $l$, which exchanges information between the two token streams. The synchronization layer $f$ computes a function  ($\mathbf{x}_{l+1}^\text{v}, \mathbf{x}_{l+1}^\text{t}) = f(\mathbf{y}_l^\text{v}, \mathbf{y}_l^\text{t})$ in order to obtain the input tokens for the next layer. This architecture is shown in Fig.~\ref{fig:overview}. 
Several model designs have been proposed to extend pre-trained video models for 4D video generation.
Next, we will review and analyze designs in previous works (Sec.~\ref{sec:seq}). Then we introduce our design of the synchronization layer (Sec.~\ref{sec:syn1}-\ref{sec:syn2}).

\subsubsection{Sequential interleaving} 
\label{sec:seq}
A competing design choice would be to compute alternating updates for temporal and multi-view consistency:
\vspace{-0.5mm}
\begin{equation}
        \mathbf{y}_l =  \mathbf{x}_l + \varphi_l^\text{v}(\mathbf{x}_l; \mathbf{c}^\text{v}),\quad
        \mathbf{x}_{l+1} =  \mathbf{y}_l + \varphi_l^\text{t}(\mathbf{y}_l; \mathbf{c}^\text{t}).
    \label{eq:seq_interleave}
    \vspace{-0.5mm}
\end{equation}
where $\varphi_l^\text{v}$ and $\varphi_l^\text{v}$ denote applying the transformer layer $\varphi_l$ across the view and the time axes, respectively. Most prior works~\cite{vividzoo,4diffusion,cvd} that sequentially interleave cross-view attention and cross-time attention, could be interpreted as performing the above steps.

While conceptually simple, this approach has limitations:
(i) It does not fully account for the interdependence between temporal and view consistency, treating them as independent objectives during each update.
(ii) Outputs from the view update $\mathbf{y}_l$ may be out-of-distribution for the temporal update, leading to artifacts or reduced generalization.
Prior works either train an additional network to adapt $\mathbf{y}_l$ to become in-distribution~\cite{vividzoo}, or fine-tune the cross-view attention or the cross-time attention to compensate for the discrepancy~\cite{4diffusion,cvd}. However, due to the limited available 4D data, fine-tuning the attention layers could degrade the quality of the video model, limiting its generalization capability to domains outside the 4D training set.

\subsubsection{Synchronization in Optimization} 
\label{sec:syn1}
One can interpret the DiT blocks of the video model as functions performing a fixed number of iterative variable updates to optimize an \emph{implicit} cost function $\mathcal{C}(\mathbf{x}; \mathbf{c})$~\cite{ahn2023transformers,jastrzkebski2017residual}.
The implicit cost function $\mathcal{C}$ can be thought of as an abstract measure of "closeness" to realistic videos given context $\mathbf{c}$. Under certain restricted assumptions, Ahn~\etal \cite{ahn2023transformers} prove that for a transformer
with $L$ layers, it learns to perform $L$ iterations of preconditioned gradient descent to reach certain critical points of the training loss. 

The intuition that the transformer architecture can be seen as an \emph{iterative optimization solver} motivates us to create a link to the optimization literature to explain our synchronization layer design choices. Using the optimization analogy, our video model solves a combined optimization problem for 4D generation:
\vspace{-0.5mm}
\begin{equation}
    \min_\mathbf{x} \mathcal{C}_\text{v}(\mathbf{x}) + \mathcal{C}_\text{t}(\mathbf{x}).
    \label{eq:4d_obj_1_simple}
\vspace{-0.5mm}
\end{equation}
where $\mathcal{C}_\text{v}$ ensures that each row of the grid is a fixed-view video and $\mathcal{C}_\text{t}$ ensures that each column is a freeze-time video.
Using the idea of variable splitting, this problem can be transformed into the equivalent problem:
\vspace{-0.5mm}
\begin{equation}
    \min_{(\mathbf{x}^\text{v}, \mathbf{x}^\text{t})}
    \mathcal{C}_\text{v}(\mathbf{x}^\text{v}) + \mathcal{C}_\text{t}(\mathbf{x}^\text{t})\quad
s.t.\, \mathbf{x}^\text{v} = \mathbf{x}^\text{t}.
    \label{eq:4d_obj_1_simple}
\vspace{-0.5mm}
\end{equation}

An optimization problem with this structure can be tackled by algorithms like projected gradient descent, which performs a projection on the constraint manifold at every iteration. This leads to the design of a hard synchronization between the two token streams. 
Alternatively, one can employ a quadratic regularization or an algorithm like ADMM~\cite{parikh2014proximal} that does not strictly enforce the constraint at every iteration but makes the token streams more similar. This leads to the design of a soft synchronization between the two token streams.


\subsubsection{Synchronization layer design}
\label{sec:syn2}

The synchronization layer maintains consistency between the two token streams, as defined in E.q.~\ref{eq:parallel}. Following this, we explore two synchronization strategies:

\textbf{Hard synchronization.}
Hard synchronization strictly enforces the constraint $\mathbf{x}_l^\text{t} = \mathbf{x}_l^\text{v}$ at every iteration. A straightforward approach to hard synchronization is to compute an update by averaging tokens.
However, in contrast to traditional optimization, we can generalize this step to compute a weighted combination with learned weights:
\vspace{-1mm}
\begin{equation}
    \mathbf{x}_{l+1} = \mathbf{W}_l^\text{v} \mathbf{y}_{l}^\text{v} + \mathbf{W}_l^\text{t} \mathbf{y}_{l}^\text{t},
    \label{eq:hard_sync}
    \vspace{-1mm}
\end{equation}
where $\mathbf{W}_l^\text{v}$, $\mathbf{W}_l^\text{t}$ are linear weights for merging each token with initial values, \ie, $\frac{1}{2} \mathbf I$. The weights are modulated by the diffusion time $\sigma$ to make them adaptive to different stages of the diffusion process.

Empirically, the 4D model with hard sync can indeed generate temporally consistent 4D videos.
However, it tends to produce less desirable frames when the viewpoint differs significantly from the input fixed-view video. Common artifacts include objects appearing stretched in the direction of camera movement or unintended object motion when the time stamp is intended to be frozen (Refer to visual examples in Fig.~\ref{fig:ablation}). We hypothesize that the limitation of hard sync is that the merged video tokens are aggregated from both the freeze-time and fixed-view videos, causing a discrepancy in the learned distribution of the base video model. 

\begin{figure}[h]
    \centering
    \begin{subfigure}{0.48\linewidth}
        \centering
        \includegraphics[width=\linewidth]{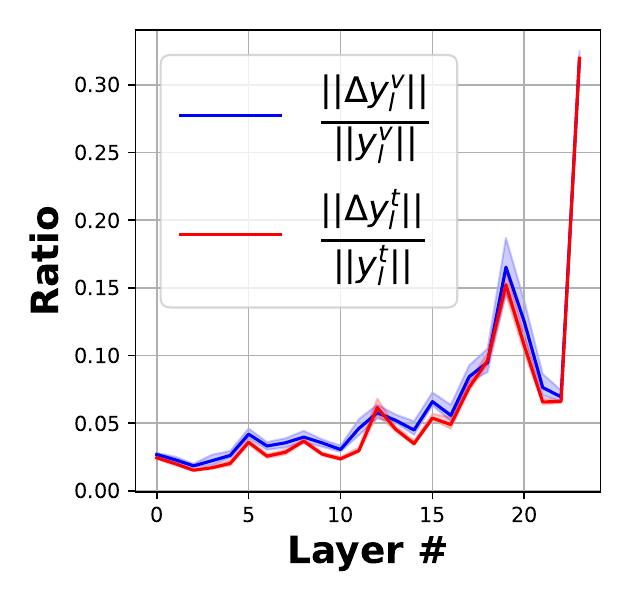}
        \caption{Relative magnitude of $\Delta \mathbf{y}_l^\text{v}$, $\Delta \mathbf{y}_l^\text{t}$ in Eq.~\eqref{eq:soft_sync} at each layer. }
    \end{subfigure}
    \hspace{0.01\linewidth} 
    \begin{subfigure}{0.48\linewidth}
        \centering
        \includegraphics[width=\linewidth]{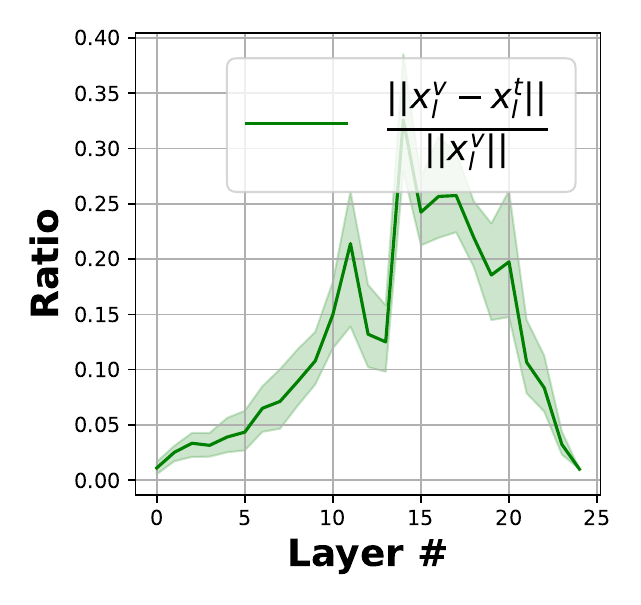}
        \caption{Similarity between $\mathbf{x}_l^\text{v}$ and $\mathbf{x}_l^\text{t}$ at each layer.}
    \end{subfigure}
    \vspace{-2mm}
    \caption{The dynamics of soft synchronization during inference.}
    \label{fig:softsync}
    \vspace{-4mm}
\end{figure}

\textbf{Soft synchronization.}
The above observation motivates an alternative soft synchronization strategy -- the video tokens $\mathbf{x}_l^\text{v}$,$\mathbf{x}_l^\text{t}$ are kept in two separate streams instead of merging them into a single copy as in Eq.~\eqref{eq:hard_sync}. A soft update is used to make the streams more similar. This design gives additional flexibility for the model to adaptively adjust the strength of synchronization at different layers. Again, we can design a more general solution as would be available in traditional optimization and use a modulated linear layer to predict asymmetrical token updates:
\vspace{-1mm}
\begin{equation}
(\Delta\mathbf{y}_l^\text{v}, \Delta\mathbf{y}_l^\text{t})  = \text{Mod\_Linear}(\mathbf{y}_l^\text{v},  \mathbf{y}_l^\text{t}; \sigma).
\vspace{-1mm}
\end{equation}
Then, the tokens are updated separately:
\vspace{-1mm}
\begin{equation}
    \mathbf{x}_{l+1}^\text{v} = \mathbf{y}_l^\text{v} + \Delta\mathbf{y}_l^\text{v},\quad
    \mathbf{x}_{l+1}^\text{t} = \mathbf{y}_l^\text{t} + \Delta\mathbf{y}_l^\text{t}.
    \label{eq:soft_sync}
\vspace{-1mm}
\end{equation}
Soft synchronization offers more flexibility, adapting the strength of synchronization across layers. Empirically, this results in better consistency and fewer artifacts in challenging scenarios, such as large viewpoint changes.
We visualize the update strength and token similarity in Fig.~\ref{fig:softsync}. We observe that the update strength increases for layers deeper in the network. The token similarity is initially drifting between the two token streams before they are made more similar by the increased update strength in later layers.

\subsection{Implementation}
\noindent\textbf{Training.} The model is trained with the velocity matching loss of rectified flow~\cite{liu2022flow}, leveraging two data sources: (1) \emph{2D transformed videos:} we apply a sequence of continuous 2D affine transformations to video frames to mimic camera motion. This provides large-scale pseudo 4D data to train the model to generate synchronized motions. However, models only trained with this source tend to generate flattened foreground objects that are noticeable when changing viewpoints. (2) \emph{Animated Objaverse dataset:} We render around 15,000 multi-view videos using animated 3D assets from Objaverse~\cite{objaverse}, positioning the rendering cameras on a circular trajectory around each asset. Fine-tuning with this small-scale, synthetic, object-centric 4D dataset quickly equips the model with the ability to maintain both temporal and multi-view consistency, even in complex scenes containing multiple objects and intricate environments.

\noindent\textbf{Extending to a wider view and longer time.} The model is trained to generate an $8\times8$ frame grid in each step. For input fixed-view videos or freeze-time videos with extended durations, we generate frames autoregressively, advancing along the time and view axes in a sliding window fashion.

\section{Experiments}
\label{sec:experiments}

\begin{figure}[t]
\begin{center}
\includegraphics[width=\linewidth]{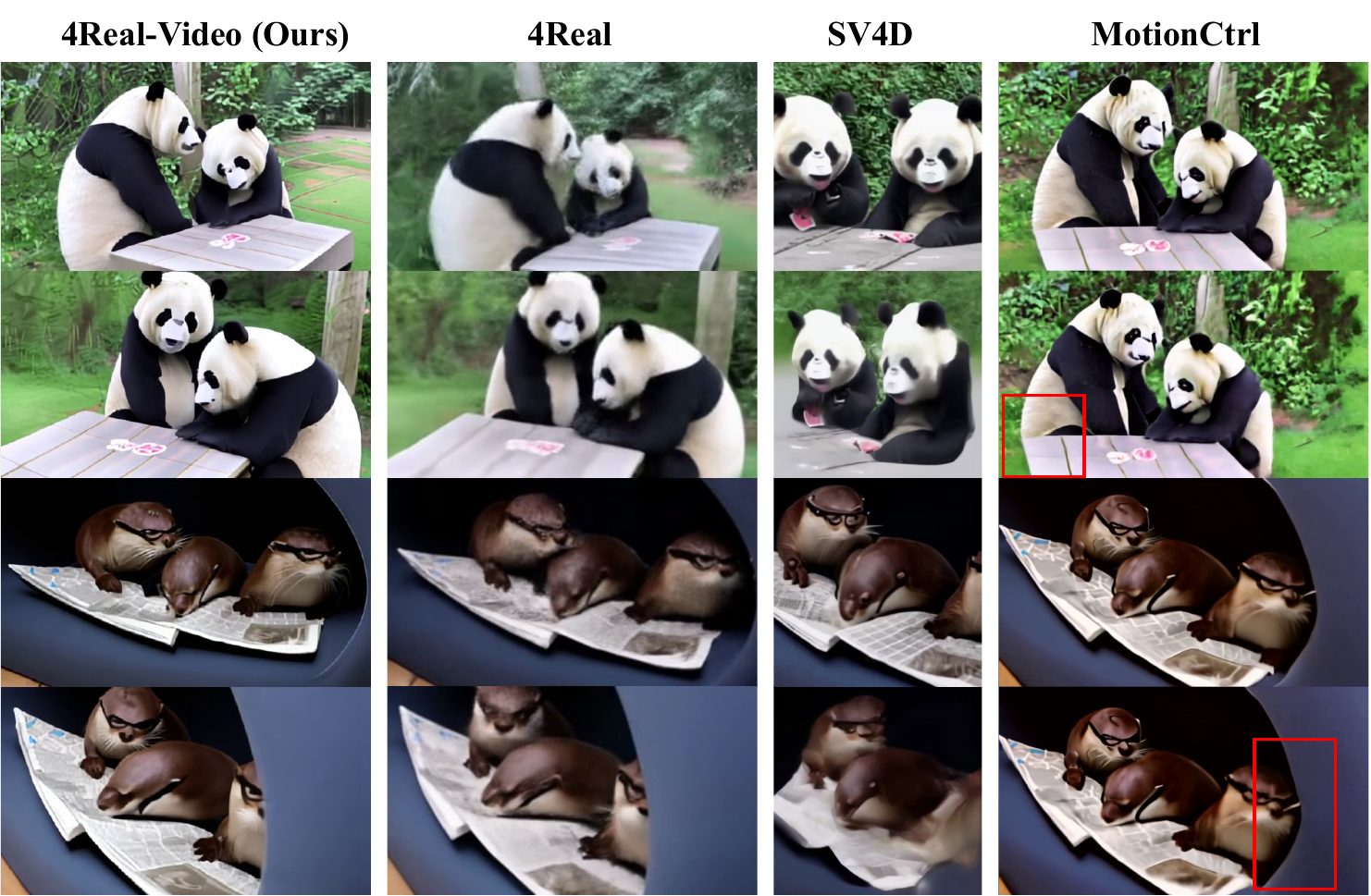}
\vspace{-2mm}
\caption{\textbf{Visual Comparisons.} We show two viewpoints for a fixed time for each method. Our method produces high-quality images, even under significant camera motion. In contrast, frames generated by 4Real and SV4D tend to appear more blurred, with objects notably distorted in SV4D. MotionCtrl struggles to generate frames under substantial camera motion. We use \textcolor{red}{red} bounding boxes to highlight regions with distortions and flickering, which become particularly noticeable when viewed as a video.
}
\label{fig:compare}
\vspace{-4mm}
\end{center}
\end{figure}

\begin{figure*}[t]
\begin{center}
\includegraphics[width=\linewidth]{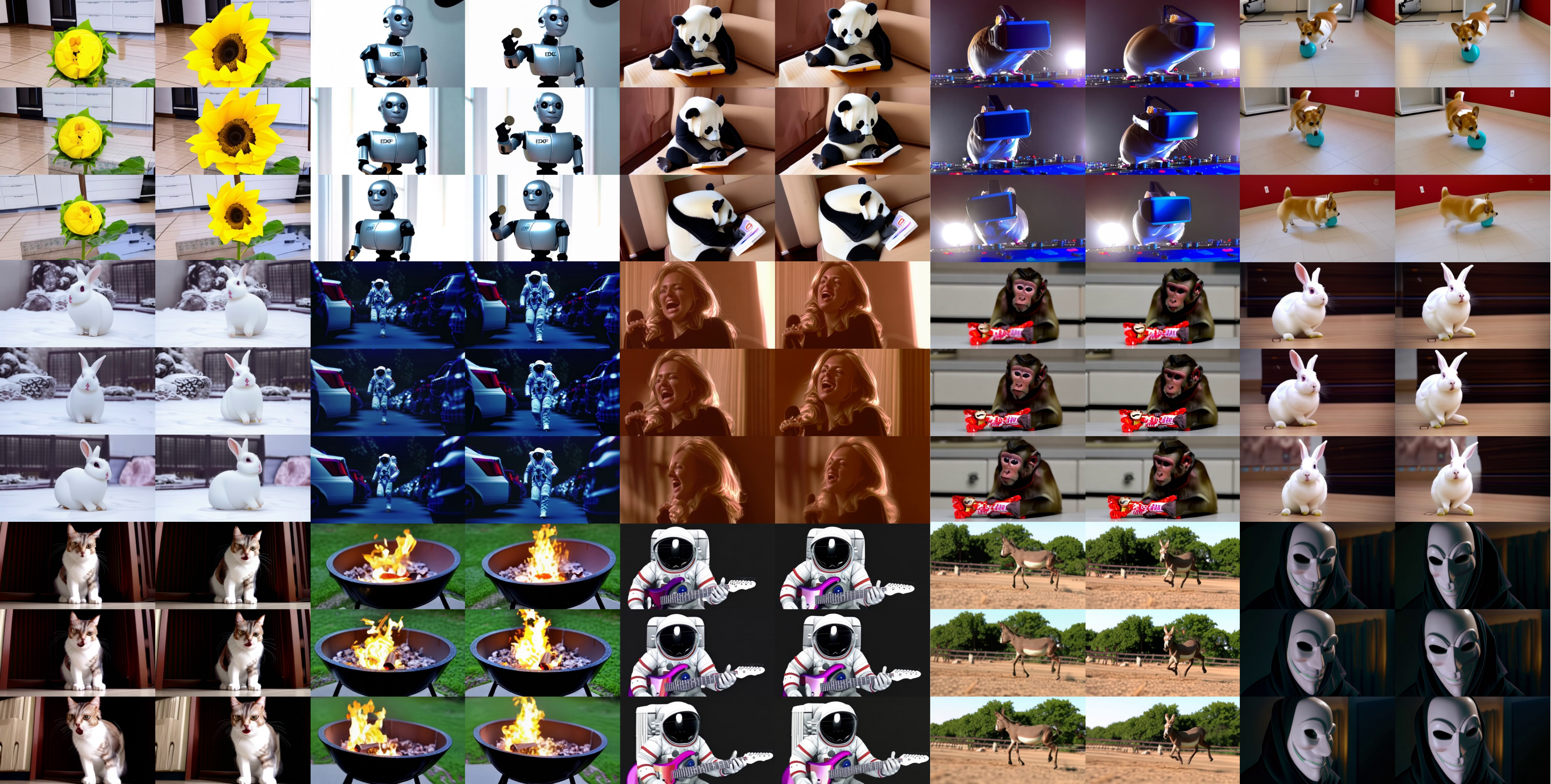}
\vspace{-5mm}
\caption{\textbf{Results from \Approach}. We can generate diverse and high-quality dynamic content.
}
\label{fig:results}
\vspace{-4mm}
\end{center}
\end{figure*}

\begin{figure*}[t]
\begin{center}
\includegraphics[width=\linewidth]{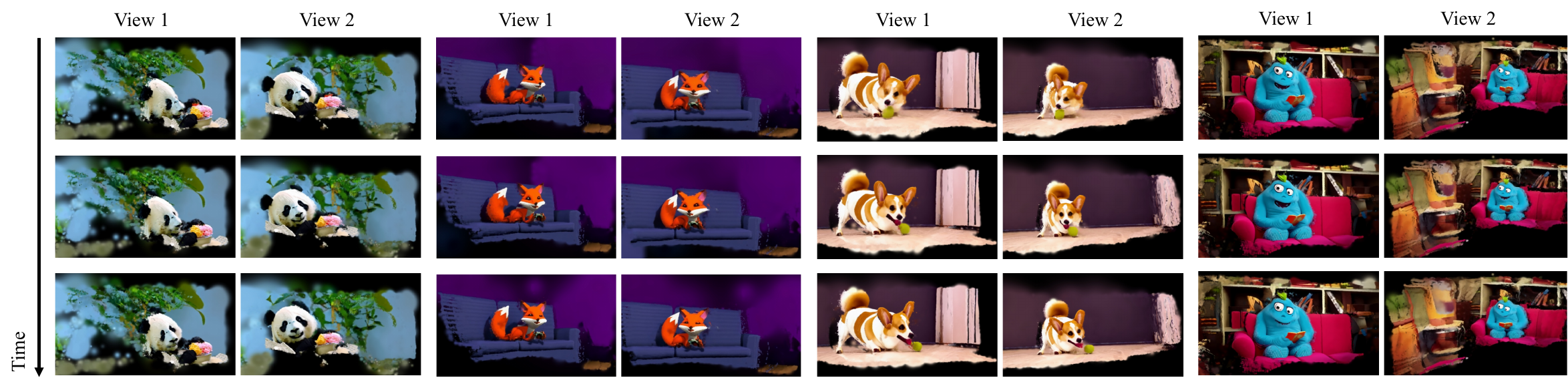}
\vspace{-4mm}
\caption{\textbf{Deformable 3D Gaussian Splatting Reconstruction} from the generated 4D videos demonstrate the spatial and temporal consistency of the proposed method. 
}
\vspace{-6mm}
\label{fig:3dgs_result}
\end{center}
\end{figure*}

\begin{table*}[h]
    \centering
     \resizebox{\textwidth}{!}{
        \begin{tabular}{l|cccccccccccc}
            \toprule
        \textbf{Method} & \textbf{FID} $\downarrow$ & \textbf{CLIP} $\uparrow$ & \multicolumn{2}{c}{\textbf{FVD} $\downarrow$} 
        & \multicolumn{2}{c}{\textbf{FVD-Test} $\downarrow$}
         & \multicolumn{2}{c}{\textbf{Visual Quality$\uparrow$}} & \multicolumn{2}{c}{\textbf{Temporal Consist.$\uparrow$}} & \multicolumn{2}{c}{\textbf{Factual Consist. $\uparrow$}}\\
            &&& Time & View & Time & View& Time & View & Time & View & Time & View \\
            \midrule
            SV4D~\cite{sv4d} &204.81 &19.46 & 1053.10 &1245.42 &814.50 & 323.99 & 2.26 & 2.02 & 2.03  & 1.68 & 2.12 & 1.99 \\
            MotionCtrl~\cite{motionctrl} 
            &87.10 &20.20 & 1556.36 & 1509.76 &1170.04 &302.18 & 2.36 & 2.30 & 2.38 & 2.25 & 2.38 & 2.33 \\
            \hline
            Sequential &96.64  &28.16 & 1662.54 & 1797.15 &897.08 &597.19 & 2.30 & 2.28 & 2.21 & 2.15 & 2.23 & 2.20 \\
            Soft w/o Obj 
            &80.17 &28.11 &1392.48
            &1720.47 & 318.18 &302.18 
             & 2.41 & 2.39 & 2.37 & 2.31 & 2.35 & 2.33 \\
            Hard Sync &79.92 &28.16 & 972.87 & 1045.35 &316.14 & \textbf{251.44}
            & 2.42 & 2.40 & 2.40 & 2.33 & 2.37 & 2.34 \\
            Soft Sync &\textbf{78.36} &\textbf{28.22}& \textbf{906.16} & \textbf{1036.00} &\textbf{308.15} &261.02  & \textbf{2.43} & \textbf{2.42} & \textbf{2.41} & \textbf{2.36} & \textbf{2.38} & \textbf{2.36} \\
            \bottomrule
        \end{tabular}
    }
    \vspace{-2mm}
    \caption{\textbf{Quantitative ablation}. We evaluate the visual quality, temporal consistency, and text-video alignment using various metrics.}
    \vspace{-3mm}
    \label{tab:ablation}
\end{table*}

\begin{table}[h]
\resizebox{0.49 \textwidth}{!}{
\begin{tabular}{l|ccc|ccc}
\toprule
 & \multicolumn{3}{c}{\textbf{GIM-Confidence$\uparrow$}} & \multicolumn{3}{c}{\textbf{Dust3R-Confidence$\uparrow$}} \\
 & $\tau=0.1$ & $\tau=0.5$ & $\tau=0.7$ & $\tau=2.0$ & $\tau=2.5$ & $\tau=3.0$\\
 \midrule
Sequential & 65.0 & 28.6 & 18.5 & 33.5 & 24.6 & 16.6 \\
Soft w/o Obj & \textbf{80.8} & \textbf{49.6} & \textbf{36.1} & 39.1 & 31.4 & 24.0\\ 
Hard Sync & 78.7 & 44.8 & 31.4 & 39.3 & 31.5 & 23.8\\
Soft Sync & 79.6 & 47.1 & 33.8 & \textbf{41.0} & \textbf{33.4} & \textbf{25.7}\\
\bottomrule
\end{tabular}
}
\vspace{-2mm}
\caption{\textbf{Multi-view consistency } is measured using an image-matching method and a 3D reconstruction method.  
}
\vspace{-3mm}
\label{tab:mv_consistency}
\end{table}

\subsection{Implementation details}
\noindent\textbf{Base video model.} The base video model consists of 600M parameters, with 24 DiT blocks of 1024 channel size. We found that pixel-based diffusion models train faster and produce more coherent motion compared to latent-based models of similar model size. Thus, we opt to train the base model to directly output pixel values, given limited accessible GPU resources. The model is progressively trained from a resolution of 36$\times$64 to 72$\times$128  using 24 A100 GPUs for 12 days. We then train a diffusion-based upsampler to upsample the video to the target 288$\times$512 resolution. 

\noindent\textbf{4D  model.} The 4D video model is trained progressively from low to high resolution. It is first trained using pseudo 4D videos for 20k iterations, followed by fine-tuning for 3k iterations on the Animated Objaverse dataset~\cite{objaverse}. Notably, longer training on the Objaverse data led to a slight decrease in quality when applied to real-world scenes. Note that fine-tuning only affects the weights of the synchronization layers to avoid shifting the video distribution away from real videos.

\subsection{Evaluation}
\noindent\textbf{Test sets.} We use the Snap Video Model~\cite{snapvideo} to generate pairs of freeze-time and fixed-view videos, given a diverse set of text prompts. Each video is 2 seconds long, consisting of 16 frames. In total, we collected 200 pairs to serve as testing inputs. Some samples of our results on the test set are shown in Fig.~\ref{fig:results}.

\noindent\textbf{Evaluation metrics.} Evaluating 4D video generation is challenging without ground truth data. We employ the following metrics to assess generation quality:

\textbullet\ \textbf{VideoScore}~\cite{videoscore} is a video quality evaluation network that outputs scores assessing visual quality, temporal consistency, text-video alignment, motion degree, and factual consistency. In our case, we drop text-video alignment and motion degree scores since these scores are more related to the input conditional videos instead of generated frames. 

\textbullet\ \textbf{FVD}~\cite{fvd} evaluates the Frechet Distance between the generated video distribution and the data distribution. 
We reported two versions of the FVD score: (1) against a large dataset of real videos, where the score is relatively high due to the distribution mismatch caused by the out-of-distribution content of our test cases, and (2) against statistics computed from the input test set videos to provide a more relevant comparison.

\textbullet\ \textbf{CLIP Score}~\cite{clip} evaluates the similarity between generated images against the text prompt. It also reflects the visual quality of the generated frames.

\textbullet\ \textbf{GIM-Confidence.} We utilize GIM~\cite{gim}, a state-of-the-art 2D image matching method, to measure the consistency of appearance across views. Specifically, we report the proportion of matching pixels across views under different confidence thresholds. Note the GIM focuses on 2D image matching and cannot reflect 3D consistency well.

\textbullet\ \textbf{Dust3R-Confidence.} To further evaluate \emph{3D} multi-view consistency, we use Dust3R~\cite{dust3r}, a state-of-the-art 3D reconstruction network to analyze generated freeze-time videos. Dust3R provides pixel-wise confidence scores reflecting 3D multi-view consistency, and we report the proportion of pixels above different confidence thresholds.
 
We evaluated \emph{VideoScore} and \emph{FVD} for both videos playing either along the time axes as fixed-view videos, or along the view axes as freeze-time videos, in order to evaluate both temporal and multi-view consistency of the generated frame grid. \emph{GIM} and \emph{Dust3R-Confidence} are used only in ablations with fixed camera trajectories, where the confidence scores are comparable.

\begin{figure*}[t]
\begin{center}
\includegraphics[width=\linewidth]{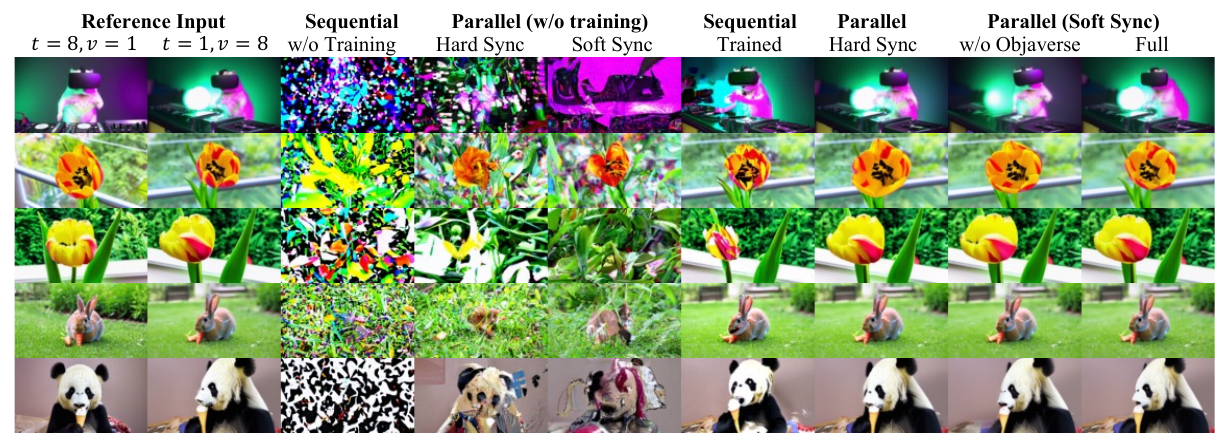}
\vspace{-6mm}
\caption{\textbf{Ablation comparisons.} We visually compare the video quality and consistency among different design choices. 
}
\label{fig:ablation}
\vspace{-6mm}
\end{center}
\end{figure*}

\begin{figure}[t]
\centering
\includegraphics[width=1\linewidth]{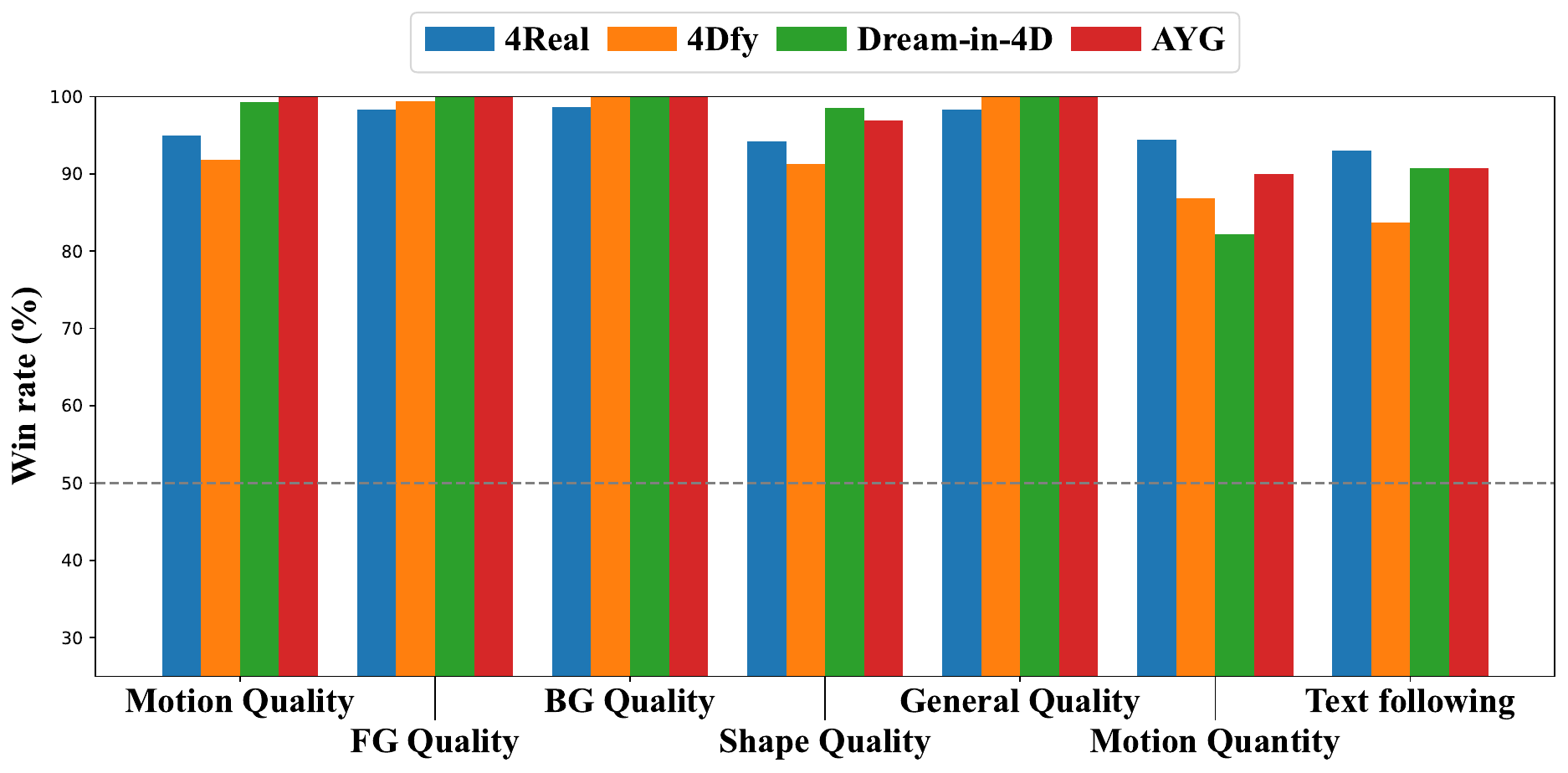}
\vspace{-6mm}
\caption{\textbf{User study} against optimization-based 4D generation methods across different rating criteria.}
\label{fig:user-study}
\vspace{-5mm}
\end{figure}

\noindent\textbf{Comparison against 4D video generation baselines.} There is currently no prior method that functions exactly like ours, so we establish two baselines:
(1) we use MotionCtrl~\cite{motionctrl}, a state-of-the-art camera control video generation method, to generate freeze-time videos for each frame of the input fixed-view videos.
The videos are generated with ``Round-RI\_90'' camera trajectory and speed parameter set to 4.0 to encourage larger camera motions.
(2) We run SV4D~\cite{sv4d}, a state-of-the-art 4D video model trained specifically for animated 3D objects.

MotionCtrl fails to generate temporally coherent videos because freeze-time videos are generated independently, ignoring temporal dependencies. It also tends to generate very small camera motion despite being given a large input speed. On the other hand, SV4D fails to create meaningful results when applied to realistic-style videos, which are out of its training domain. In comparison, our method generates realistic and coherent frame grids and achieves higher scores across different metrics, as shown in Table~\ref{tab:ablation}.

\noindent\textbf{Comparison against optimization-based 4D generation baselines.} We also compare against recent 4D generation methods~\cite{4dfy,4real,zheng2024unified,ayg} that rely on computationally expensive score distillation sampling~\cite{dreamfusion}. Due to the limited number of samples we can acquire from these methods, and the fact that these samples were generated using different settings (object-centric v.s. scene-level), we conducted a user study instead.

The study involves 10 evaluators per video pair. In each session, evaluators were presented with two anonymized videos. Each video depicted a dynamic object or scene, with the camera moving along a circular trajectory and stopping at 2-4 poses to highlight object motions. We obtained 16 videos for 4Dfy~\cite{4dfy}, 14 videos for Dream-in-4D~\cite{zheng2024unified}, 14 videos for AYG~\cite{ayg} and 36 videos for 4Real~\cite{4real} from their respective project web pages. Evaluators were tasked with selecting their preferences based on seven criteria: \emph{motion realism}, \emph{foreground/background quality}, \emph{shape realism}, \emph{general quality}, \emph{motion quality}, and \emph{video-text alignment}. As shown in Figure~\ref{fig:user-study}, our method outperformed the competition in every category by a large margin. More details of the user study are provided in the supplementary.



\subsection{Ablations}
We analyze our method by comparing it against the following variations: (1) a sequential interleaved architecture (see Eq.~\eqref{eq:seq_interleave}); (2) training only with a pseudo-4D video dataset without Objaverse; (3) using hard synchronization; and (4) our full method with soft synchronization. 
The results are shown in Table~\ref{tab:ablation}, Table~\ref{tab:mv_consistency} and visualized in Fig.~\ref{fig:ablation}. Further details of each ablated design are provided in the supplementary material. 
We make the following observations: First, the proposed parallel architecture achieves better performance compared to the sequential architecture. Second, training our model without any 4D data can still produce competitive results compared to baselines, showing the robustness of our approach. It obtains higher GIM-Confidence as the metric favors only image matching instead of real 3D consistency. Finally, soft synchronization improves quality over hard synchronization, leading to more coherent and visually appealing outputs.

\subsection{Deformable 3D Gaussian Splatting reconstruction from generated 4D videos} To further validate the effectiveness of our method in generating explicit 3D representations, we fit deformable 3D Gaussian Splatting (3DGS) to the generated 4D videos. Fig.~\ref{fig:3dgs_result} qualitatively shows reconstructed 3DGS at different times and viewpoints. More details of the reconstruction pipeline are included in the supplementary.


\section{Conclusion}
\label{sec:conclusion}


We propose \Approach, a novel framework for 4D video generation. The core idea of our framework is to process a grid of frames using two separate token streams that are processed in parallel, with a synchronization layer coordinating between the two streams. Remarkably, our model can generate diverse photorealistic 4D videos without requiring access to such a dataset.
Despite its strengths, our current implementation has several \textbf{limitations} that we aim to address in future work. First, the base video model’s small size constrains its capability, limiting the visual quality and resolution of the generated videos. This can be improved by incorporating more advanced and larger-scale video models. Second, our framework currently lacks support for 360$^\circ$ video generation. Enhancing this capability will involve improving the training of the base video model and incorporating camera pose conditioning. Third, generating freeze-time videos remains a significant challenge, particularly for dynamic elements such as running horses or fires, where robustness is limited. Finally, our approach requires post-processing steps to construct explicit 3D representations of the generated dynamic scenes. In the future, it would be exciting to explore the possibility of a single feedforward model for 4D generation and further advancing the field.


\clearpage
{
    \small
    \bibliographystyle{ieeenat_fullname}
    \bibliography{main}

\begin{thebibliography}{54}
\providecommand{\natexlab}[1]{#1}
\providecommand{\url}[1]{\texttt{#1}}
\expandafter\ifx\csname urlstyle\endcsname\relax
  \providecommand{\doi}[1]{doi: #1}\else
  \providecommand{\doi}{doi: \begingroup \urlstyle{rm}\Url}\fi

\bibitem[Ahn et~al.(2023)Ahn, Cheng, Daneshmand, and Sra]{ahn2023transformers}
Kwangjun Ahn, Xiang Cheng, Hadi Daneshmand, and Suvrit Sra.
\newblock Transformers learn to implement preconditioned gradient descent for in-context learning.
\newblock \emph{NeurIPS}, 2023.

\bibitem[Bahmani et~al.(2024{\natexlab{a}})Bahmani, Skorokhodov, Rong, Wetzstein, Guibas, Wonka, Tulyakov, Park, Tagliasacchi, and Lindell]{4dfy}
Sherwin Bahmani, Ivan Skorokhodov, Victor Rong, Gordon Wetzstein, Leonidas Guibas, Peter Wonka, Sergey Tulyakov, Jeong~Joon Park, Andrea Tagliasacchi, and David~B Lindell.
\newblock 4d-fy: Text-to-4d generation using hybrid score distillation sampling.
\newblock In \emph{CVPR}, 2024{\natexlab{a}}.

\bibitem[Bahmani et~al.(2024{\natexlab{b}})Bahmani, Skorokhodov, Siarohin, Menapace, Qian, Vasilkovsky, Lee, Wang, Zou, Tagliasacchi, et~al.]{vd3d}
Sherwin Bahmani, Ivan Skorokhodov, Aliaksandr Siarohin, Willi Menapace, Guocheng Qian, Michael Vasilkovsky, Hsin-Ying Lee, Chaoyang Wang, Jiaxu Zou, Andrea Tagliasacchi, et~al.
\newblock Vd3d: Taming large video diffusion transformers for 3d camera control.
\newblock \emph{arXiv preprint arXiv:2407.12781}, 2024{\natexlab{b}}.

\bibitem[Bain et~al.(2021)Bain, Nagrani, Varol, and Zisserman]{webvid10m}
Max Bain, Arsha Nagrani, G{\"u}l Varol, and Andrew Zisserman.
\newblock Frozen in time: A joint video and image encoder for end-to-end retrieval.
\newblock In \emph{ICCV}, 2021.

\bibitem[Chen et~al.(2023{\natexlab{a}})Chen, Xia, He, Zhang, Cun, Yang, Xing, Liu, Chen, Wang, Weng, and Shan]{videocrafter}
Haoxin Chen, Menghan Xia, Yingqing He, Yong Zhang, Xiaodong Cun, Shaoshu Yang, Jinbo Xing, Yaofang Liu, Qifeng Chen, Xintao Wang, Chao Weng, and Ying Shan.
\newblock Videocrafter1: Open diffusion models for high-quality video generation.
\newblock \emph{arXiv preprint arXiv:2310.19512}, 2023{\natexlab{a}}.

\bibitem[Chen et~al.(2023{\natexlab{b}})Chen, Chen, Jiao, and Jia]{fantasia3d}
Rui Chen, Yongwei Chen, Ningxin Jiao, and Kui Jia.
\newblock Fantasia3d: Disentangling geometry and appearance for high-quality text-to-3d content creation.
\newblock In \emph{ICCV}, 2023{\natexlab{b}}.

\bibitem[Deitke et~al.(2023)Deitke, Schwenk, Salvador, Weihs, Michel, VanderBilt, Schmidt, Ehsani, Kembhavi, and Farhadi]{objaverse}
Matt Deitke, Dustin Schwenk, Jordi Salvador, Luca Weihs, Oscar Michel, Eli VanderBilt, Ludwig Schmidt, Kiana Ehsani, Aniruddha Kembhavi, and Ali Farhadi.
\newblock Objaverse: A universe of annotated 3d objects.
\newblock In \emph{Proceedings of the IEEE/CVF Conference on Computer Vision and Pattern Recognition}, pages 13142--13153, 2023.

\bibitem[Guo et~al.(2023)Guo, Yang, Rao, Liang, Wang, Qiao, Agrawala, Lin, and Dai]{animatediff}
Yuwei Guo, Ceyuan Yang, Anyi Rao, Zhengyang Liang, Yaohui Wang, Yu Qiao, Maneesh Agrawala, Dahua Lin, and Bo Dai.
\newblock Animatediff: Animate your personalized text-to-image diffusion models without specific tuning.
\newblock \emph{arXiv preprint arXiv:2307.04725}, 2023.

\bibitem[He et~al.(2024{\natexlab{a}})He, Xu, Guo, Wetzstein, Dai, Li, and Yang]{cameractrl}
Hao He, Yinghao Xu, Yuwei Guo, Gordon Wetzstein, Bo Dai, Hongsheng Li, and Ceyuan Yang.
\newblock Cameractrl: Enabling camera control for text-to-video generation.
\newblock \emph{arXiv preprint arXiv:2404.02101}, 2024{\natexlab{a}}.

\bibitem[He et~al.(2024{\natexlab{b}})He, Jiang, Zhang, Ku, Soni, Siu, Chen, Chandra, Jiang, Arulraj, et~al.]{videoscore}
Xuan He, Dongfu Jiang, Ge Zhang, Max Ku, Achint Soni, Sherman Siu, Haonan Chen, Abhranil Chandra, Ziyan Jiang, Aaran Arulraj, et~al.
\newblock Mantisscore: Building automatic metrics to simulate fine-grained human feedback for video generation.
\newblock \emph{arXiv preprint arXiv:2406.15252}, 2024{\natexlab{b}}.

\bibitem[Hessel et~al.(2021)Hessel, Holtzman, Forbes, Bras, and Choi]{clip}
Jack Hessel, Ari Holtzman, Maxwell Forbes, Ronan~Le Bras, and Yejin Choi.
\newblock Clipscore: {A} reference-free evaluation metric for image captioning.
\newblock \emph{CoRR}, abs/2104.08718, 2021.

\bibitem[Ho et~al.(2022)Ho, Chan, Saharia, Whang, Gao, Gritsenko, Kingma, Poole, Norouzi, Fleet, et~al.]{imagenvideo}
Jonathan Ho, William Chan, Chitwan Saharia, Jay Whang, Ruiqi Gao, Alexey Gritsenko, Diederik~P Kingma, Ben Poole, Mohammad Norouzi, David~J Fleet, et~al.
\newblock Imagen video: High definition video generation with diffusion models.
\newblock \emph{arXiv preprint arXiv:2210.02303}, 2022.

\bibitem[Jastrzębski et~al.(2017)Jastrzębski, Arpit, Ballas, Verma, Che, and Bengio]{jastrzkebski2017residual}
Stanis{\l}aw Jastrzębski, Devansh Arpit, Nicolas Ballas, Vikas Verma, Tong Che, and Yoshua Bengio.
\newblock Residual connections encourage iterative inference.
\newblock \emph{arXiv preprint arXiv:1710.04773}, 2017.

\bibitem[Jiang et~al.(2023)Jiang, Zhang, Gao, Hu, and Yao]{consistent4d}
Yanqin Jiang, Li Zhang, Jin Gao, Weimin Hu, and Yao Yao.
\newblock Consistent4d: Consistent 360 $\{$$\backslash$deg$\}$ dynamic object generation from monocular video.
\newblock \emph{arXiv preprint arXiv:2311.02848}, 2023.

\bibitem[Kerbl et~al.(2023)Kerbl, Kopanas, Leimk{\"u}hler, and Drettakis]{3dgs}
Bernhard Kerbl, Georgios Kopanas, Thomas Leimk{\"u}hler, and George Drettakis.
\newblock 3d gaussian splatting for real-time radiance field rendering.
\newblock \emph{ACM Trans. Graph.}, 42\penalty0 (4):\penalty0 139--1, 2023.

\bibitem[Kuang et~al.(2024)Kuang, Cai, He, Xu, Li, Guibas, and Wetzstein]{cvd}
Zhengfei Kuang, Shengqu Cai, Hao He, Yinghao Xu, Hongsheng Li, Leonidas Guibas, and Gordon. Wetzstein.
\newblock Collaborative video diffusion: Consistent multi-video generation with camera control.
\newblock In \emph{arXiv}, 2024.

\bibitem[Li et~al.(2024)Li, Zheng, Zhu, Mai, Zhang, Wonka, and Ghanem]{vividzoo}
Bing Li, Cheng Zheng, Wenxuan Zhu, Jinjie Mai, Biao Zhang, Peter Wonka, and Bernard Ghanem.
\newblock Vivid-zoo: Multi-view video generation with diffusion model, 2024.

\bibitem[Lin et~al.(2023)Lin, Gao, Tang, Takikawa, Zeng, Huang, Kreis, Fidler, Liu, and Lin]{magic3d}
Chen-Hsuan Lin, Jun Gao, Luming Tang, Towaki Takikawa, Xiaohui Zeng, Xun Huang, Karsten Kreis, Sanja Fidler, Ming-Yu Liu, and Tsung-Yi Lin.
\newblock Magic3d: High-resolution text-to-3d content creation.
\newblock In \emph{CVPR}, 2023.

\bibitem[Ling et~al.(2024)Ling, Kim, Torralba, Fidler, and Kreis]{ayg}
Huan Ling, Seung~Wook Kim, Antonio Torralba, Sanja Fidler, and Karsten Kreis.
\newblock Align your gaussians: Text-to-4d with dynamic 3d gaussians and composed diffusion models.
\newblock In \emph{CVPR}, 2024.

\bibitem[Liu et~al.(2023)Liu, Wu, Van~Hoorick, Tokmakov, Zakharov, and Vondrick]{zero123}
Ruoshi Liu, Rundi Wu, Basile Van~Hoorick, Pavel Tokmakov, Sergey Zakharov, and Carl Vondrick.
\newblock Zero-1-to-3: Zero-shot one image to 3d object.
\newblock In \emph{ICCV}, 2023.

\bibitem[Liu et~al.(2022)Liu, Gong, and Liu]{liu2022flow}
Xingchao Liu, Chengyue Gong, and Qiang Liu.
\newblock Flow straight and fast: Learning to generate and transfer data with rectified flow.
\newblock \emph{arXiv preprint arXiv:2209.03003}, 2022.

\bibitem[Menapace et~al.(2024)Menapace, Siarohin, Skorokhodov, Deyneka, Chen, Kag, Fang, Stoliar, Ricci, Ren, et~al.]{snapvideo}
Willi Menapace, Aliaksandr Siarohin, Ivan Skorokhodov, Ekaterina Deyneka, Tsai-Shien Chen, Anil Kag, Yuwei Fang, Aleksei Stoliar, Elisa Ricci, Jian Ren, et~al.
\newblock Snap video: Scaled spatiotemporal transformers for text-to-video synthesis.
\newblock In \emph{CVPR}, 2024.

\bibitem[Ngo et~al.(2024)Ngo, Zhuang, Gan, Kalogerakis, Tulyakov, Lee, and Wang]{delta}
Tuan~Duc Ngo, Peiye Zhuang, Chuang Gan, Evangelos Kalogerakis, Sergey Tulyakov, Hsin-Ying Lee, and Chaoyang Wang.
\newblock Delta: Dense efficient long-range 3d tracking for any video, 2024.

\bibitem[{OpenAI}(2024)]{sora}
{OpenAI}.
\newblock Video generation models as world simulators, 2024.
\newblock Accessed: 2024-11-08.

\bibitem[Parikh et~al.(2014)Parikh, Boyd, et~al.]{parikh2014proximal}
Neal Parikh, Stephen Boyd, et~al.
\newblock Proximal algorithms.
\newblock \emph{Foundations and trends{\textregistered} in Optimization}, 2014.

\bibitem[Peebles and Xie(2022)]{Peebles2022DiT}
William Peebles and Saining Xie.
\newblock Scalable diffusion models with transformers.
\newblock \emph{arXiv preprint arXiv:2212.09748}, 2022.

\bibitem[Polyak et~al.(2024)Polyak, Zohar, Brown, Tjandra, Sinha, Lee, Vyas, Shi, Ma, Chuang, Yan, Choudhary, Wang, Sethi, Pang, Ma, Misra, Hou, Wang, Jagadeesh, Li, Zhang, Singh, Williamson, Le, Yu, Singh, Zhang, Vajda, Duval, Girdhar, Sumbaly, Rambhatla, Tsai, Azadi, Datta, Chen, Bell, Ramaswamy, Sheynin, Bhattacharya, Motwani, Xu, Li, Hou, Hsu, Yin, Dai, Taigman, Luo, Liu, Wu, Zhao, Kirstain, He, He, Pumarola, Thabet, Sanakoyeu, Mallya, Guo, Araya, Kerr, Wood, Liu, Peng, Vengertsev, Schonfeld, Blanchard, Juefei-Xu, Nord, Liang, Hoffman, Kohler, Fire, Sivakumar, Chen, Yu, Gao, Georgopoulos, Moritz, Sampson, Li, Parmeggiani, Fine, Fowler, Petrovic, and Du]{moviegen}
Adam Polyak, Amit Zohar, Andrew Brown, Andros Tjandra, Animesh Sinha, Ann Lee, Apoorv Vyas, Bowen Shi, Chih-Yao Ma, Ching-Yao Chuang, David Yan, Dhruv Choudhary, Dingkang Wang, Geet Sethi, Guan Pang, Haoyu Ma, Ishan Misra, Ji Hou, Jialiang Wang, Kiran Jagadeesh, Kunpeng Li, Luxin Zhang, Mannat Singh, Mary Williamson, Matt Le, Matthew Yu, Mitesh~Kumar Singh, Peizhao Zhang, Peter Vajda, Quentin Duval, Rohit Girdhar, Roshan Sumbaly, Sai~Saketh Rambhatla, Sam Tsai, Samaneh Azadi, Samyak Datta, Sanyuan Chen, Sean Bell, Sharadh Ramaswamy, Shelly Sheynin, Siddharth Bhattacharya, Simran Motwani, Tao Xu, Tianhe Li, Tingbo Hou, Wei-Ning Hsu, Xi Yin, Xiaoliang Dai, Yaniv Taigman, Yaqiao Luo, Yen-Cheng Liu, Yi-Chiao Wu, Yue Zhao, Yuval Kirstain, Zecheng He, Zijian He, Albert Pumarola, Ali Thabet, Artsiom Sanakoyeu, Arun Mallya, Baishan Guo, Boris Araya, Breena Kerr, Carleigh Wood, Ce Liu, Cen Peng, Dimitry Vengertsev, Edgar Schonfeld, Elliot Blanchard, Felix Juefei-Xu, Fraylie Nord, Jeff Liang, John Hoffman, Jonas
  Kohler, Kaolin Fire, Karthik Sivakumar, Lawrence Chen, Licheng Yu, Luya Gao, Markos Georgopoulos, Rashel Moritz, Sara~K. Sampson, Shikai Li, Simone Parmeggiani, Steve Fine, Tara Fowler, Vladan Petrovic, and Yuming Du.
\newblock Movie gen: A cast of media foundation models, 2024.

\bibitem[Poole et~al.(2023)Poole, Jain, Barron, and Mildenhall]{dreamfusion}
Ben Poole, Ajay Jain, Jonathan~T Barron, and Ben Mildenhall.
\newblock Dreamfusion: Text-to-3d using 2d diffusion.
\newblock In \emph{ICLR}, 2023.

\bibitem[Ren et~al.(2023)Ren, Pan, Tang, Zhang, Cao, Zeng, and Liu]{dreamgaussian4d}
Jiawei Ren, Liang Pan, Jiaxiang Tang, Chi Zhang, Ang Cao, Gang Zeng, and Ziwei Liu.
\newblock Dreamgaussian4d: Generative 4d gaussian splatting.
\newblock \emph{arXiv preprint arXiv:2312.17142}, 2023.

\bibitem[Rombach et~al.(2022)Rombach, Blattmann, Lorenz, Esser, and Ommer]{ldm}
Robin Rombach, Andreas Blattmann, Dominik Lorenz, Patrick Esser, and Bj{\"o}rn Ommer.
\newblock High-resolution image synthesis with latent diffusion models.
\newblock In \emph{CVPR}, 2022.

\bibitem[Saharia et~al.(2022)Saharia, Chan, Saxena, Li, Whang, Denton, Ghasemipour, Gontijo~Lopes, Karagol~Ayan, Salimans, et~al.]{imagen}
Chitwan Saharia, William Chan, Saurabh Saxena, Lala Li, Jay Whang, Emily~L Denton, Kamyar Ghasemipour, Raphael Gontijo~Lopes, Burcu Karagol~Ayan, Tim Salimans, et~al.
\newblock Photorealistic text-to-image diffusion models with deep language understanding.
\newblock \emph{NeurIPS}, 2022.

\bibitem[Shao et~al.(2024)Shao, Pang, Zheng, Sun, and Liu]{human4dit}
Ruizhi Shao, Youxin Pang, Zerong Zheng, Jingxiang Sun, and Yebin Liu.
\newblock Human4dit: Free-view human video generation with 4d diffusion transformer.
\newblock \emph{arXiv preprint arXiv:2405.17405}, 2024.

\bibitem[Shen et~al.(2024)Shen, Cai, Yin, M{\"u}ller, Li, Wang, Chen, and Wang]{gim}
Xuelun Shen, Zhipeng Cai, Wei Yin, Matthias M{\"u}ller, Zijun Li, Kaixuan Wang, Xiaozhi Chen, and Cheng Wang.
\newblock Gim: Learning generalizable image matcher from internet videos.
\newblock \emph{arXiv preprint arXiv:2402.11095}, 2024.

\bibitem[Shi et~al.(2024)Shi, Wang, Ye, Long, Li, and Yang]{mvdream}
Yichun Shi, Peng Wang, Jianglong Ye, Mai Long, Kejie Li, and Xiao Yang.
\newblock Mvdream: Multi-view diffusion for 3d generation.
\newblock In \emph{ICLR}, 2024.

\bibitem[Singer et~al.(2023)Singer, Sheynin, Polyak, Ashual, Makarov, Kokkinos, Goyal, Vedaldi, Parikh, Johnson, et~al.]{mav3d}
Uriel Singer, Shelly Sheynin, Adam Polyak, Oron Ashual, Iurii Makarov, Filippos Kokkinos, Naman Goyal, Andrea Vedaldi, Devi Parikh, Justin Johnson, et~al.
\newblock Text-to-4d dynamic scene generation.
\newblock \emph{arXiv preprint arXiv:2301.11280}, 2023.

\bibitem[Smart et~al.(2024)Smart, Zheng, Laina, and Prisacariu]{splatt3r}
Brandon Smart, Chuanxia Zheng, Iro Laina, and Victor~Adrian Prisacariu.
\newblock Splatt3r: Zero-shot gaussian splatting from uncalibrated image pairs.
\newblock 2024.

\bibitem[Tucker and Snavely(2018)]{realestate10k}
Richard Tucker and Noah Snavely.
\newblock Stereo magnification: Learning view synthesis using multiplane images.
\newblock In \emph{ACM TOG}, 2018.

\bibitem[Unterthiner et~al.(2018)Unterthiner, van Steenkiste, Kurach, Marinier, Michalski, and Gelly]{fvd}
Thomas Unterthiner, Sjoerd van Steenkiste, Karol Kurach, Rapha{\"e}l Marinier, Marcin Michalski, and Sylvain Gelly.
\newblock Fvd: A new metric for video generation.
\newblock \emph{arXiv preprint arXiv:1812.01717}, 2018.

\bibitem[Wang et~al.(2023{\natexlab{a}})Wang, Du, Li, Yeh, and Shakhnarovich]{sjc}
Haochen Wang, Xiaodan Du, Jiahao Li, Raymond~A Yeh, and Greg Shakhnarovich.
\newblock Score jacobian chaining: Lifting pretrained 2d diffusion models for 3d generation.
\newblock In \emph{CVPR}, 2023{\natexlab{a}}.

\bibitem[Wang et~al.(2024{\natexlab{a}})Wang, Leroy, Cabon, Chidlovskii, and Revaud]{dust3r}
Shuzhe Wang, Vincent Leroy, Yohann Cabon, Boris Chidlovskii, and Jerome Revaud.
\newblock Dust3r: Geometric 3d vision made easy.
\newblock In \emph{CVPR}, 2024{\natexlab{a}}.

\bibitem[Wang et~al.(2023{\natexlab{b}})Wang, Lu, Wang, Bao, Li, Su, and Zhu]{prolificdreamer}
Zhengyi Wang, Cheng Lu, Yikai Wang, Fan Bao, Chongxuan Li, Hang Su, and Jun Zhu.
\newblock Prolificdreamer: High-fidelity and diverse text-to-3d generation with variational score distillation.
\newblock In \emph{NeurIPS}, 2023{\natexlab{b}}.

\bibitem[Wang et~al.(2024{\natexlab{b}})Wang, Yuan, Wang, Li, Chen, Xia, Luo, and Shan]{motionctrl}
Zhouxia Wang, Ziyang Yuan, Xintao Wang, Yaowei Li, Tianshui Chen, Menghan Xia, Ping Luo, and Ying Shan.
\newblock Motionctrl: A unified and flexible motion controller for video generation.
\newblock In \emph{ACM SIGGRAPH}, 2024{\natexlab{b}}.

\bibitem[Watson et~al.(2024)Watson, Saxena, Li, Tagliasacchi, and Fleet]{4DiM}
Daniel Watson, Saurabh Saxena, Lala Li, Andrea Tagliasacchi, and David~J Fleet.
\newblock Controlling space and time with diffusion models.
\newblock \emph{arXiv preprint arXiv:2407.07860}, 2024.

\bibitem[Xie et~al.(2024)Xie, Yao, Voleti, Jiang, and Jampani]{sv4d}
Yiming Xie, Chun-Han Yao, Vikram Voleti, Huaizu Jiang, and Varun Jampani.
\newblock Sv4d: Dynamic 3d content generation with multi-frame and multi-view consistency.
\newblock \emph{arXiv preprint arXiv:2407.17470}, 2024.

\bibitem[Xu et~al.(2024)Xu, Nie, Liu, Liu, Kautz, Wang, and Vahdat]{camco}
Dejia Xu, Weili Nie, Chao Liu, Sifei Liu, Jan Kautz, Zhangyang Wang, and Arash Vahdat.
\newblock Camco: Camera-controllable 3d-consistent image-to-video generation.
\newblock \emph{arXiv preprint arXiv:2406.02509}, 2024.

\bibitem[Yang et~al.(2024{\natexlab{a}})Yang, Hou, Huang, Ma, Wan, Zhang, Chen, and Liao]{directavideo}
Shiyuan Yang, Liang Hou, Haibin Huang, Chongyang Ma, Pengfei Wan, Di Zhang, Xiaodong Chen, and Jing Liao.
\newblock Direct-a-video: Customized video generation with user-directed camera movement and object motion.
\newblock In \emph{ACM SIGGRAPH}, 2024{\natexlab{a}}.

\bibitem[Yang et~al.(2024{\natexlab{b}})Yang, Teng, Zheng, Ding, Huang, Xu, Yang, Hong, Zhang, Feng, et~al.]{cogvideox}
Zhuoyi Yang, Jiayan Teng, Wendi Zheng, Ming Ding, Shiyu Huang, Jiazheng Xu, Yuanming Yang, Wenyi Hong, Xiaohan Zhang, Guanyu Feng, et~al.
\newblock Cogvideox: Text-to-video diffusion models with an expert transformer.
\newblock \emph{arXiv preprint arXiv:2408.06072}, 2024{\natexlab{b}}.

\bibitem[Yin et~al.(2023)Yin, Xu, Wang, Zhao, and Wei]{4dgen}
Yuyang Yin, Dejia Xu, Zhangyang Wang, Yao Zhao, and Yunchao Wei.
\newblock 4dgen: Grounded 4d content generation with spatial-temporal consistency.
\newblock \emph{arXiv preprint arXiv:2312.17225}, 2023.

\bibitem[Yu et~al.(2024)Yu, Wang, Zhuang, Menapace, Siarohin, Cao, Jeni, Tulyakov, and Lee]{4real}
Heng Yu, Chaoyang Wang, Peiye Zhuang, Willi Menapace, Aliaksandr Siarohin, Junli Cao, Laszlo~A Jeni, Sergey Tulyakov, and Hsin-Ying Lee.
\newblock 4real: Towards photorealistic 4d scene generation via video diffusion models.
\newblock In \emph{NeurIPS}, 2024.

\bibitem[Zhang et~al.(2024)Zhang, Chen, Wang, Liu, Wang, and Qiao]{4diffusion}
Haiyu Zhang, Xinyuan Chen, Yaohui Wang, Xihui Liu, Yunhong Wang, and Yu Qiao.
\newblock 4diffusion: Multi-view video diffusion model for 4d generation.
\newblock \emph{arXiv preprint arXiv:2405.20674}, 2024.

\bibitem[Zhao et~al.(2023)Zhao, Yan, Xie, Hong, Li, and Lee]{animate124}
Yuyang Zhao, Zhiwen Yan, Enze Xie, Lanqing Hong, Zhenguo Li, and Gim~Hee Lee.
\newblock Animate124: Animating one image to 4d dynamic scene.
\newblock \emph{arXiv preprint arXiv:2311.14603}, 2023.

\bibitem[Zhao et~al.(2024)Zhao, Lin, Lin, Yan, Li, Yang, Wang, Lee, and Wang]{genxd}
Yuyang Zhao, Chung-Ching Lin, Kevin Lin, Zhiwen Yan, Linjie Li, Zhengyuan Yang, Jianfeng Wang, Gim~Hee Lee, and Lijuan Wang.
\newblock Genxd: Generating any 3d and 4d scenes.
\newblock \emph{arXiv preprint arXiv:2411.02319}, 2024.

\bibitem[Zheng et~al.(2024)Zheng, Li, Nagano, Liu, Hilliges, and Mello]{zheng2024unified}
Yufeng Zheng, Xueting Li, Koki Nagano, Sifei Liu, Otmar Hilliges, and Shalini~De Mello.
\newblock A unified approach for text- and image-guided 4d scene generation.
\newblock In \emph{CVPR}, 2024.

\bibitem[Zhu and Zhuang(2023)]{hifa}
Joseph Zhu and Peiye Zhuang.
\newblock Hifa: High-fidelity text-to-3d with advanced diffusion guidance.
\newblock In \emph{ICLR}, 2023.

\end{thebibliography}
}

\clearpage
\setcounter{page}{1}
\maketitlesupplementary


\section{Deformable 3D GS 
reconstruction details}
Using generated 4D videos with multi-view frame grids, we apply a reconstruction method to produce an explicit 3D representation, \ie, deformable 3D geometric structures (GS).

\noindent\textbf{Canonical 3D representation.} We use 3D Gaussian Splats~\cite{3dgs} to represent the canonical shape of the dynamic scene. This representation consists of a set of 3D Gaussian points defined by their 3D position, orientation, scale, opacity, and RGB color. The 3D Gaussian Splats are rendered by projecting the Gaussian points onto the image plane and aggregating pixel values using a NeRF-like volumetric rendering equation. In our implementation, we find that constraining the Gaussians to be isotropic effectively reduces artifacts when viewing the 3D representation from viewpoints distinct from the training perspectives.

\noindent\textbf{Deformation field.} To model a 4D scene, we use a deformation field to represent the offsets of the 3DGS. This deformation field is implemented as an MLP, which takes the 3D position of a point and time as input and outputs a 3D displacement offset.

\noindent\textbf{Initialize canonical 3D GS with 3D dense tracking.} While the input freeze-time video may appear visually plausible, it is not truly geometrically accurate, particularly in the background regions. Directly optimizing 3D GS using these frames as ground truth results in significant artifacts. The most noticeable issue is the noisy reconstruction of background regions, which fail to separate cleanly from the foreground. In our preliminary exploration, we tested state-of-the-art feedforward reconstruction methods, including Dust3R~\cite{dust3r} and Splatt3R~\cite{splatt3r}. However, in most cases, only the foreground regions could be reliably reconstructed, while the background remained noisy and entangled with the foreground. We attribute this limitation to the quality of the video model used to generate the inputs. In the long term, this issue could potentially be addressed by employing a higher-quality video model. At this stage, we instead use a recent 3D dense tracking method~\cite{delta}, which performs pixel-wise tracking to aggregate 3D points from various keyframes of the freeze-time video. These points are aligned towards a central frame, whose coordinates are treated as the canonical frame. 

The advantage of switching to 3D tracking is that it does not require the scene to be static, allowing it to handle multi-view inconsistencies in the generated videos by treating them as non-rigid deformations. Furthermore, 3D tracking leverages monocular depth estimation as input, preserving the clean foreground/background separation provided by the estimated depth map. This results in a visually more coherent and appealing outcome.

\noindent\textbf{Removing boundary floaters.} 3D tracking often produces outlier points along depth boundaries, a common artifact in monocular depth estimation. To eliminate these 'floaters,' we apply a rendering loss to optimize the opacity of each point, effectively pruning points that cause visual artifacts.

Specifically, given a set of aggregated 3D points from dense tracking, we know each point's 3D position in the frame coordinates of each frame of the input freeze-time video. This allows us to use the differentiable 3DGS renderer to re-render each input frame and compute the loss. Furthermore, since the points are modeled as isotropic Gaussians without orientation and are already in the frame coordinate system, we avoid the need to estimate camera extrinsics at this stage. This approach enhances robustness against multi-view inconsistencies in the input video.
 
\noindent\textbf{Temporal deformation with view-dependent compensation.} The next step involves fitting a temporal deformation field to animate the canonical 3DGS to follow the motion in the input 4D video. However, due to imperfections in the multi-view consistency of the 4D video—an issue inherited from the input freeze-time video—directly optimizing the temporal deformation field would lead to noisy reconstructions, mirroring the challenges previously discussed.

To address this issue, we augment the temporal deformation with additional view-dependent deformation to compensate for inconsistencies in the generated frames across different views. Specifically, to re-render a point on the input frame $\mathcal{I}_{ij}$ of the input frame grid, where $i$ and $j$ represent the indices of view and time respectively, the deformation offset $\Delta\mathbf{p}_{ij}$ for each point $\mathbf{p}$ in canonical space is now computed as:
\begin{equation}
    \Delta \mathbf{p}_{ij} = \Delta \mathbf{p}^\text{v}_i + \Delta \mathbf{p}^\text{t}_j,
\end{equation} 
where $\Delta \mathbf{p}^\text{t}_j$ represents the temporal deformation computed by an MLP, and $\Delta \mathbf{p}^\text{v}_i$ is the view-dependent deformation estimated via dense 3D tracking during the canonical 3DGS reconstruction stage. It is worth noting that view-dependent deformation has also been employed in 4Real~\cite{4real}; however, in our approach, the view-dependent deformation is predicted from dense 3D tracking rather than optimized using rendering loss, making it more robust.

\section{Implementation details of ablation study}
We compared different baseline variants to analyze our approach. Below, we provide details for each method corresponding to the columns in Fig.~\ref{fig:ablation}.

\noindent\textbf{Sequential w/o training.} We sequentially interleave cross-view and cross-time attention, as described in Equation \eqref{eq:seq_interleave}. All parameters in the attention layers are directly inherited from the base video model without additional training. We observe that this variant produces noisy outputs lacking meaningful structure.

\noindent\textbf{Parallel w/o training, hard sync.} We perform inference using the proposed architecture without training the synchronization layers. For hard synchronization, we average the token updates, \ie 
\begin{equation}
    \mathbf{x}_{l+1} = \frac{1}{2}(\mathbf{y}_l^\text{v} + \mathbf{y}_l^\text{t}).
\end{equation}
This version generates some content with elements from the input video, but it remains highly noisy.

\noindent\textbf{Parallel w/o training, soft sync.} The soft synchronization is implemented as weighted averaging,
\begin{equation}
    \begin{aligned}
        \mathbf{x}_{l+1}^\text{v} & = (1-w_l) \mathbf{y}_l^\text{v} + w_l \mathbf{y}_l^\text{t} \\
        \mathbf{x}_{l+1}^\text{t} & = (1-w_l) \mathbf{y}_l^\text{t} + w_l \mathbf{y}_l^\text{v}
    \end{aligned}
\end{equation}
Here, $w_l$ represents the weight, which gradually increases with the layer depth, specifically defined as $w_l = 0.1 + \frac{l}{L} \cdot 0.4$. This approach produces results with more discernible content compared to hard synchronization.

\noindent\textbf{Sequential trained.} The sequential architecture is trained following the same procedure as our proposed approach. We experimented with two variants: finetuning only the cross-time attention and finetuning only the temporal attention. Our findings indicate that finetuning temporal attention results in more stable outcomes. Therefore, for brevity, we report results only for the version where cross-time attention is finetuned.

\noindent\textbf{Parallel hard sync.} 
The variant of our proposed method employing hard synchronization.

\noindent\textbf{Parallel soft sync w/o Objaverse.} A variant of our proposed method with soft synchronization, without fine-tuning on animated 4D Objaverse data.

\section{User study details}
\begin{figure*}
    \centering
    \includegraphics[width=\linewidth]{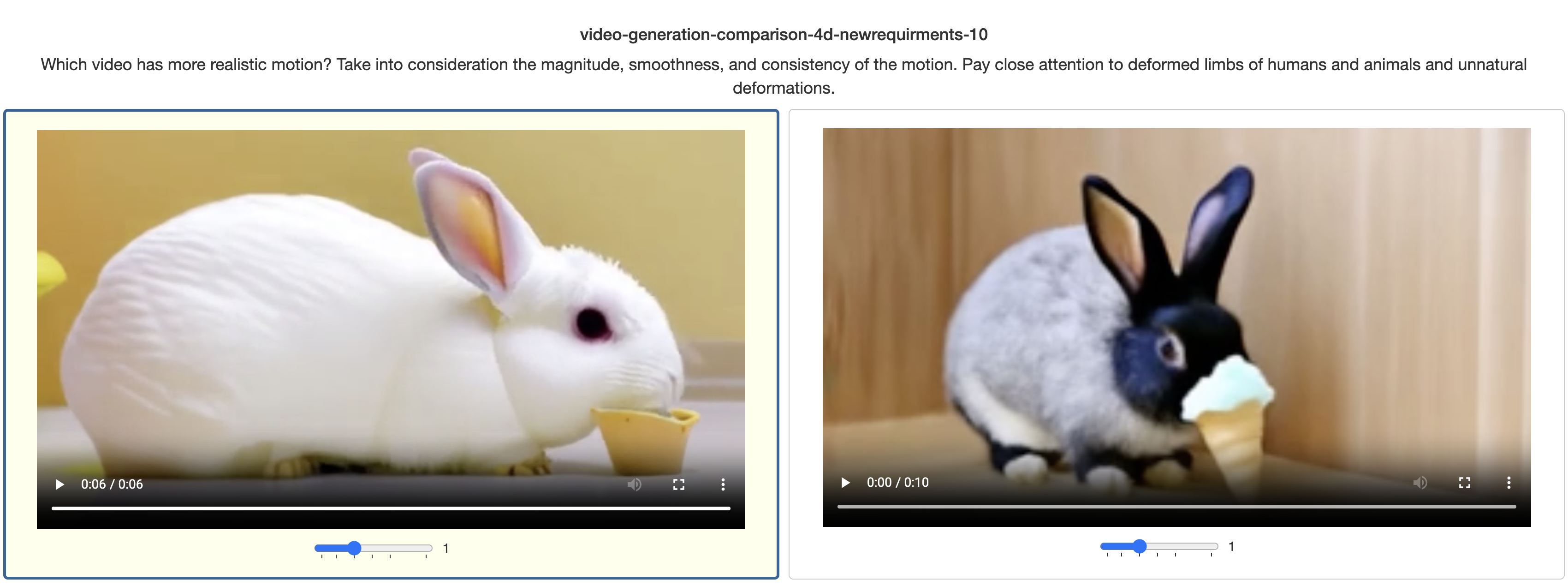}
    \caption{A screenshot of the interface for user study.}
    \label{fig:user_study_screenshot}
\end{figure*}

The user study shown in Fig.~\ref{fig:user-study} is conducted with 10 evaluators per video pair. During each session, evaluators were presented with two anonymized videos with an interface as shown in Fig.~\ref{fig:user_study_screenshot}.
The evaluators were given the following instructions:
\begin{quote}
    You are shown a description of a video and two different 3D videos generated by AI based on this description. Your task is to answer 7 questions regarding the quality of these videos. Please pay close attention to instructions and answer as thoughtfully as you can. The video shows several consecutive views of the same dynamic object. 
\begin{enumerate}
    \item  Which video has more realistic motion? Take into consideration the magnitude, smoothness, and consistency of the motion. Pay close attention to the deformed limbs of humans and animals and unnatural deformations.
    \item  Which video has the highest quality foreground? 
    \item Which video has the highest quality background? 
    \item Which video has an object of a better, more realist shape? That is the video in which the main object has the most natural shape, again paying attention to deformed limbs of humans and animals and unnatural deformations.
    \item In general which video looks higher quality?
    \item Which video is most dynamic? The video that contains the most motion. Please keep in mind that these is several views of the same dynamic video, played one after the other, so ignore all camera movement and focus solely on object movement. Please exclude from consideration any random limb deformations.
    \item Which video is better following the text description? That is which video reflects all the aspects included in the text description
    \end{enumerate}

Finally, if there is no significant difference in your opinion send the video to junk.
\end{quote}

\end{document}